\pdfoutput=1

\documentclass[11pt]{article}
\usepackage{url}
\usepackage{graphicx}
\usepackage{wrapfig}
\usepackage{booktabs}
\usepackage{longtable}
\usepackage{tcolorbox}
\usepackage{stackengine}
\usepackage{multirow}
\usepackage{amsmath}
\usepackage{pifont}
\usepackage{floatflt}
\usepackage{algorithm}
\usepackage{algpseudocode}
\usepackage{amsmath}
\usepackage{amssymb}
\usepackage{booktabs,tabularx}

\usepackage{tcolorbox}

\usepackage[]{ACL2023}
\usepackage{times}
\usepackage{latexsym}
\usepackage{amsmath}
\usepackage{graphicx}
\usepackage{booktabs}
\usepackage{bbm}
\usepackage{subcaption}
\usepackage{tabularx,ragged2e}
\usepackage{multicol,multirow}
\usepackage{enumitem}
\usepackage{soul}
\usepackage{tcolorbox}
\usepackage[toc,page]{appendix}

\usepackage[utf8]{inputenc}
\usepackage{array}
\usepackage{makecell}

\usepackage[T1]{fontenc}

\usepackage[utf8]{inputenc}

\usepackage{microtype}

\usepackage{inconsolata}

\title{Insights into Alignment: \\ Evaluating DPO and its Variants Across Multiple Tasks}

\author{Amir Saeidi \quad Shivanshu Verma \quad Md Nayem Uddin \quad Chitta Baral\\
  Arizona State University \\
  \texttt{\{ssaeidi1, sverma76, muddin11, cbaral\}@asu.edu}
}

\begin{document}
\maketitle
\begin{abstract}
This study evaluates Direct Preference Optimization (DPO) and its variants for aligning Large Language Models (LLMs) with human preferences, testing three configurations: (1) with Supervised Fine-Tuning (SFT), (2) without SFT, and (3) without SFT but using an instruction-tuned model. We further investigate how training set size influences model performance. Our evaluation spans 13 benchmarks—covering dialogue, reasoning, mathematical problem-solving, question answering, truthfulness, MT-Bench, Big Bench, and the Open LLM Leaderboard. We find that: (1) alignment methods often achieve near-optimal performance even with smaller subsets of training data; (2) although they offer limited improvements on complex reasoning tasks, they enhance mathematical problem-solving; and (3) using an instruction-tuned model improves truthfulness. These insights highlight the conditions under which alignment methods excel, as well as their limitations.

\end{abstract}

\section{Introduction}
Large Language Models (LLMs) demonstrate exceptional capabilities across various tasks, 
but aligning them with human preferences presents challenges, 
including high data demands and inconsistent performance across tasks. 
These models excel in mathematical reasoning problem-solving~\cite{ps1, ps2, ps3}, 
code generation programming~\cite{prog2, prog3, prog1}, 
text generation~\cite{tg1, tg2}, summarization, and creative writing, among other tasks. 
Notably, LLMs have achieved significant performance with human preferences, 
based on alignment methods including Supervised Fine-Tuning (SFT) and 
Reinforcement Learning from Human Feedback (RLHF)~\cite{sanh2022multitask, ouyang2022training}. 
While RLHF exhibits remarkable performance compared to just SFT, 
it faces limitations such as reward hacking \cite{liu2024statistical}. 
Therefore, Direct Preference Optimization (DPO) \cite{rafailov2023direct}, a state-of-the-art offline reinforcement learning method, has been proposed to optimize human preferences without the need for the RL process.

\begin{figure}[t!]
    \centering
    \includegraphics[width=7.5cm]{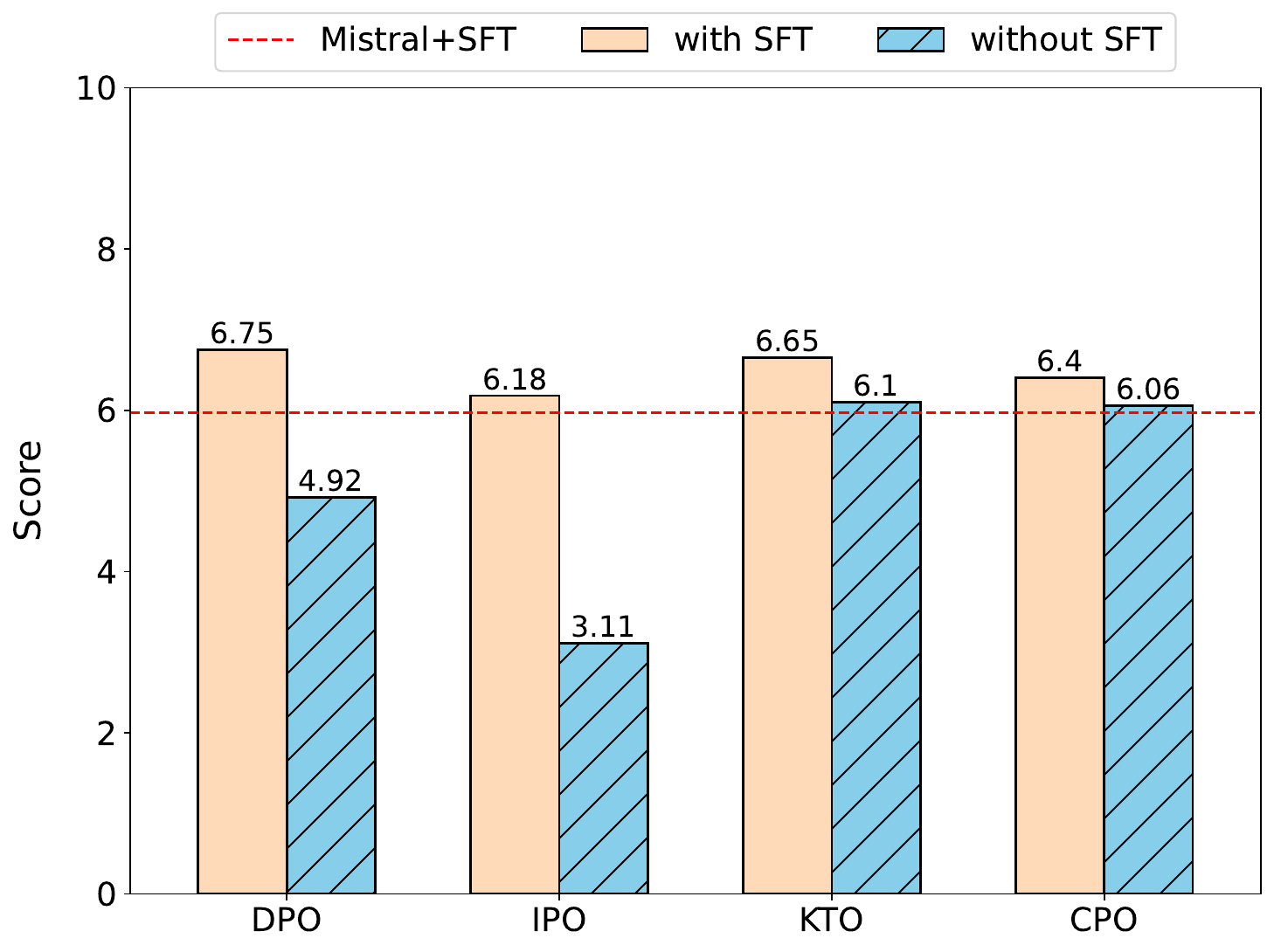}
    \caption{Performance comparison of alignment methods on MT-Bench under two scenarios: 1) fine-tuning a model with SFT (Mistral+SFT) and 2) fine-tuning a pre-trained model without SFT (Mistral). Unlike IPO and DPO, other methods like CPO and KTO demonstrate similar performance to model that undergo SFT.}
    \label{fig:mt_bench_with_without}
\end{figure}

Recent studies have highlighted limitations in alignment methods, including issues like overfitting, inefficient learning and memory utilization, preferences ranking, and dependence on preferences 
across various scenarios like dialogue systems~\cite{tunstall2023zephyr}, 
summarization, sentiment analysis~\cite{wu2023pairwise}, 
helpful and harmful question answering~\cite{liu2024statistical}, 
and machine translation~\cite{xu2024contrastive}. 
Despite the significance of these studies, 
none have thoroughly examined critical ambiguities in alignment, 
such as (1) the learnability of emerged alignment methods without SFT, 
(2) fair comparison between these methods, 
(3) evaluating their performance post-SFT, 
(4) the impact of data volume on performance, and weaknesses inherent in these methods. 
Addressing these areas is crucial for gaining a comprehensive understanding for alignment methods. 


In this study, we delve into the performance of alignment methods such as DPO~\cite{rafailov2023direct}, IPO\cite{azar2023general}, KTO~\cite{ethayarajh2023halos}, and CPO~\cite{xu2024contrastive}, which are based on RL-free algorithms. These methods typically involve two steps: 1) Supervised fine-tuning of a policy model and 2) Optimization of the SFT model with alignment algorithms such as DPO. Our exploration spans across various tasks including dialogue systems, reasoning, mathematical problem-solving, question answering, truthfulness, and multi-task understanding. We evaluate these alignment methods across 13 benchmarks such as MT-Bench~\cite{zheng2023judging}, Big Bench~\cite{srivastava2023beyond}, and Open LLM Leaderboard~\cite{open-llm-leaderboard}. To assess the performance of these methods, we define three distinct scenarios: \textbf{1) Fine-tuning an SFT model}, \textbf{2) Fine-tuning a pre-trained model}, and \textbf{3) Fine-tuning an instruction model}. In scenario 1, we employ a supervised fine-tuned model on chat completion and fine-tune it with different alignment methods. In scenario 2, we omit the SFT phase and directly fine-tune a pre-trained model with alignment methods. In scenario 3, we skip the SFT phase and utilize an instruction-tuned model as the base model, fine-tuning it with alignment methods.

The results indicate that in the standard alignment process, KTO outperforms other methods across all tasks except for multi-task understanding. However, the performance of SFT and other alignment methods in reasoning tasks is relatively comparable, suggesting that RL-free algorithms do not significantly affect reasoning. Moreover, unlike DPO when skipping the SFT phase, KTO, and CPO demonstrate comparable performance on MT-Bench. Comparing the performance of methods with and without the SFT phase reveals a significant improvement in TruthfulQA~\cite{lin2022truthfulqa} and GSM8K~\cite{cobbe2021training}. Additionally, an interesting finding is that alignment methods in the standard process exhibit better performance with smaller training data subsets. Lastly, it is observed that the instruction-tuned model has a notable impact only on truthfulness.

In summary, our contributions are as follows:
\begin{enumerate}[noitemsep]

    \item We explore the learning capabilities of alignment methods, aiming to mitigate overfitting challenges within the DPO framework. Our findings indicate that CPO and KTO show comparable performance with skipping the SFT part in MT-Bench (See Figure \ref{fig:mt_bench_with_without}).

    \item We examine the effectiveness of alignment methods across dialogue systems, reasoning, mathematical problem-solving, question answering, truthfulness, and multi-task understanding in three different scenarios.

    \item A comprehensive evaluation reveals that alignment methods exhibit a lack of performance in reasoning tasks yet demonstrate impressive performance in solving mathematical problems and truthfulness.

    \item We observe that in the standard alignment process, fine-tuning an SFT model with all alignment algorithms using a small subset of training data yields better performance. (See Figure \ref{fig:training_size}).
 
\end{enumerate}


\section{Related Works}
Recent advancements in pre-training LLMs, such as LLaMA-2 \cite{tg2}, GPT-3 \cite{gpt3}, Gopher \cite{rae2022scaling}, Vicunna \cite{vicuna2023}, Mistral \cite{jiang2023mistral}, and PaLM 2 \cite{anil2023palm}, have led to impressive performance gains in zero-shot \cite{radford2019language} and few-shot \cite{chowdhery2022palm} scenarios across various tasks. However, when applied to downstream tasks, LLMs' performance tends to degrade. While fine-tuning models using human completions aids in alignment and performance enhancement, obtaining human preferences for responses is often more feasible than collecting expert demonstrations. Consequently, recent research has shifted focus towards fine-tuning LLMs using human preferences. In this section, we present a brief review of alignment algorithms on various tasks.

RLHF \cite{christiano2023deep} proposed to optimize for maximum reward operates by engaging with a reward model trained using the Bradley-Terry (BT) model \cite{bong2022generalized} through reinforcement algorithms like Proximal Policy Optimization (PPO) \cite{schulman2017proximal}. While RLHF enhances model performance, it grapples with challenges such as instability, reward hacking, and scalability inherent in reinforcement learning. 

Recent studies have introduced methods to address these challenges by optimizing relative preferences without depending on reinforcement learning (RL). Optimizing a model using the BT model on preference datasets helps ensure alignment with human preferences.

Sequence Likelihood Calibration (SLiC) \cite{zhao2023slichf} introduced a novel approach to ranking preferences produced by a supervised fine-tuned (SFT) model, employing calibration loss and regularization fine-tuning loss during training. Meanwhile, Rank Response with Human Feedback (RRHF) \cite{yuan2023rrhf} trains the SFT model utilizing a zero-margin likelihood contrastive loss, assuming multiple ranked responses for each input. Despite their efficacy, SLiC and RRHF lack theoretical underpinnings. In response, DPO proposed a method to fit an SFT model directly to human preferences using the Bradley-Terry (BT) model, offering theoretical insights into the process.

Statistical Rejection Sampling Optimization (RSO) \cite{liu2024statistical} combines the methodologies of SLiC and DPO while introducing an enhanced method for gathering preference pairs through statistical rejection sampling. IPO \cite{azar2023general}, akin to DPO approaches, has mathematically demonstrated the limitations of the DPO approach regarding overfitting and generalization, proposing a comprehensive objective for learning from human preferences. Zephyr \cite{tunstall2023zephyr} has enhanced DPO by leveraging state-of-the-art (SOTA) models to generate responses for the same input and ranking them using teacher models like GPT-4. Additionally, they highlight the necessity of SFT as a preliminary step before employing DPO.

KTO \cite{ethayarajh2023halos}, inspired by Kahneman and Tversky’s seminal work on prospect theory \cite{KahnemanandTversky}, aims to maximize the utility of LLM generations directly rather than maximizing the log-likelihood of preferences. This approach eliminates the need for two preferences for the same input, as it focuses on discerning whether a preference is desirable or undesirable.

Self-Play fIne-tuNing (SPIN) \cite{chen2024selfplay} introduced a self-training approach to enhance DPO using the dataset employed in the SFT step. The key idea of this approach is to utilize synthetic data generated as the rejected response and the gold response from the SFT dataset as the chosen response. Meanwhile, Constrictive Preference Optimization (CPO) \cite{xu2024contrastive} proposed an efficient method for learning preferences by combining the maximum-likelihood loss and the DPO loss function, aiming to improve memory and learning efficiency.


We note that the aforementioned works lack comparative studies on alignment methods concerning both completion and preference learning. While those studies address unlearning a DPO method without the SFT step, further exploration of alternative methods is warranted. Although the significance of high-quality preferences is widely acknowledged, there remains a necessity to explore the influence of data quantity on performance of the alignment methods. Additionally, the crucial aspect of generalization remains unexplored. While aligning a model aims to enhance performance across all categories, improving alignment methods often comes at the expense of performance in other areas. Further investigation in this regard is necessary. To this end, we examine the performance of alignment methods both before and after SFT to assess the learning capabilities of IPO, KTO, and CPO. Moreover, we highlight the weaknesses of alignment methods by comparing their performance across five different domains, demonstrating the significant impact of dataset quantity on performance.

\begin{figure*}[t!]
    \centering
    \includegraphics[width=15.9cm]{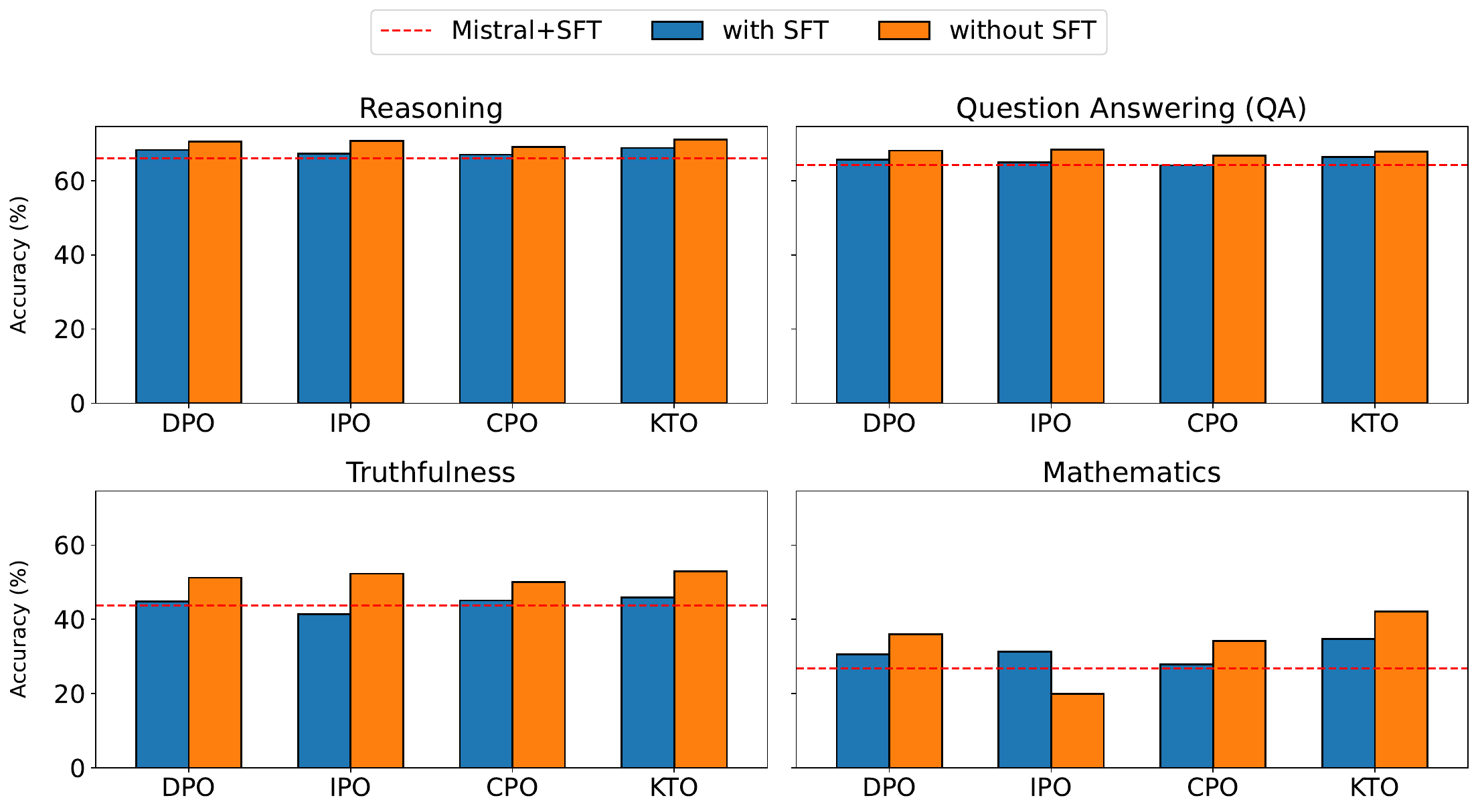}
    \caption{Comparison performance of the alignment method in different tasks based on two different scenarios: 1) fine-tuning an SFT model (Mistral+SFT) with alignment methods and 2)  fine-tuning a pre-train model (Mistral) with them. For more details about reasoning and question answering, refer to Appendix \ref{sec:more_sc1_sc2}.}
    \label{fig:comparision_wo}
\end{figure*}

\section{Exiting Alignment Methods}
In this section, we explain various RL-free alignment methods and discuss the reasons behind their development. Typically, the alignment process unfolds in three phases: 1) Fine-tuning a policy model using Supervised Fine-Tuning (SFT), 2) training a reward model, and 3) further fine-tuning the initial policy model using reinforcement learning (RL), where the reward model provides the feedback mechanism. A recent development by DPO introduced an RL-free approach aimed at aligning a policy model by optimizing the likelihood of the preferred and unpreferred responses. This is implemented using a dataset labeled $D$, where $x$ represents the input, $y_{w}$ denotes the preferred response, and $y_{l}$ indicates the unpreferred response. The DPO loss function is mathematically articulated in Equation \ref{dpo_loss} as follows:
\begin{equation}
\resizebox{\linewidth}{!}{
\begin{math}
\begin{aligned}
\mathcal{L}_{\mathrm{DPO}}\left(\pi_\theta ; \pi_{\text {ref }}\right)= & -\mathbb{E}_{\left(x, y_w, y_l\right) \sim \mathcal{D}}\left[\operatorname { l o g } \sigma \left(\beta \log \frac{\pi_\theta\left(y_w \mid x\right)}{\pi_{\text {ref }}\left(y_w \mid x\right)}\right.\right. \\
& \left.\left.-\beta \log \frac{\pi_\theta\left(y_l \mid x\right)}{\pi_{\text {ref }}\left(y_l \mid x\right)}\right)\right]
\label{dpo_loss}
\end{aligned}
\end{math}}
\end{equation}

where $\pi_{\theta}$ is the parameterized policy, $\sigma$ is sigmoid function and $\beta$ is a parameter controlling the deviation from the base reference policy $\pi_{\text{ref}}$. Despite DPO surpassing RLHF through RL-free methodology, it faces constraints like overfitting and the need for extensive regularization, which can impede the efficacy of the policy model. Addressing these limitations, in ~\cite{azar2023general} introduced the IPO algorithm, which defines a general form of the DPO and reformulates it to solve the overfitting and regularization. The formulation of the IPO loss function is in Equation \ref{ipo_loss} as follows:
\begin{equation}
\begin{split}
    \mathcal{L_{\mathrm{IPO}}}(\pi)= \underset{(y_w, y_l, x) \sim \mathcal{D}}{-\mathbb{E}}  \left( h_\pi(y_w, y_l, x)-\frac{ \tau^{-1}}{2} \right)^{2} 
\end{split}
\label{ipo_loss}
\end{equation}
\begin{equation*}
\begin{aligned}
    h_\pi(y, y', x)= \log \left( \frac{\pi\left(y \mid x\right)\pi_{\text{ref}}\left(y' \mid x\right)}{\pi\left(y' \mid x\right)\pi_{\text{ref}}\left(y \mid x\right)}\right) \\
\end{aligned}
\end{equation*}

where $x$ represents the input, $y_{w}$ denotes the preferred response, $y_{l}$ indicates the unpreferred response, $\pi_{\text{ref}}$ is the reference policy and $\tau$ is a real positive regularisation parameter.
Although the IPO algorithm overcomes the problems of overfitting and the need for extensive regularization present in DPO, the approach of aligning based on two preferences has different complications. The KTO study seeks to enhance the effectiveness of the DPO method by implementing a strategy that utilizes only a single preference. This method is inspired by the Kahneman \& Tversky theory, which observes that humans are more acutely affected by losses than gains of comparable magnitude. In this algorithm, having a clear understanding of whether a preference is suitable or unsuitable is crucial, eliminating the necessity for an alternative preference. The KTO loss function is defined in Equation \ref{kto_loss} as follows: 

\begin{equation}
\begin{aligned}
    \mathcal{L}_{\text{KTO}}(\pi_{\theta}, \pi_{\text{ref}}; \beta) = 
    \mathbb{E}_{x, y \sim \mathcal{D}} \left[ 1 - \hat{h}(x, y; \beta) \right] 
\label{kto_loss}
\end{aligned}
\newline
\end{equation}
\begin{equation*}
\resizebox{\linewidth-5pt}{!}{
\begin{math}
\begin{aligned}
\hat{h}(x, y; \beta)=
\begin{cases} 
\sigma\left(\beta \log \frac{ \pi_{\theta}(y|x)}{\pi_{\text{ref}}(y|x)} - \mathbb{E}_{x' \sim \mathcal{D}} \left[ \beta \text{KL}(\pi_{\theta} \parallel \pi_{\text{ref}}) \right]\right) \\ \hfill  \text{if } y \sim y_{\text{desirable}} | x, \\
\sigma\left( \mathbb{E}_{x' \sim \mathcal{D}} \left[ \beta \text{KL}(\pi_{\theta} \parallel \pi_{\text{ref}})\right]-\beta \log \frac{ \pi_{\theta}(y|x)}{\pi_{\text{ref}}(y|x)} \right),\\ \hfill \text{if } y \sim y_{\text{undesirable}} | x 
\end{cases}
\end{aligned}
\end{math}
}
\end{equation*}

where $\pi_\theta$ is the model we are optimizing, $\beta$ is a parameter controlling the deviation from the base reference policy $\pi_\text{ref}$, $\sigma$ is the logistic function, KL is the KL-divergence between the two distributions and $x$ is the input. IPO and KTO have enhanced the performance of the DPO model and addressed some of its shortcomings. However, the simultaneous loading of two models has led to inefficient learning in DPO algorithm. To improve upon this, the CPO method was developed, enhancing the efficiency of the DPO approach. Research detailed in~\cite{xu2024contrastive} demonstrated that it is unnecessary to load a reference policy model ($\pi_{ref}$) during training. By omitting the reference model from the memory, CPO increases operational efficiency, enabling the training of larger models at reduced costs compared to DPO. The CPO loss function is specified in Equation \ref{cpo_loss} as follows:
\begin{equation*}
\centering
\begin{aligned}
    \mathcal{L}_{\mathrm{NLL}}=-\mathbb{E}_{\left(x, y_w\right) \sim \mathcal{D}}\left[\log \pi_\theta\left(y_w \mid x\right)\right]
\end{aligned}
\end{equation*}
\begin{equation*}
\begin{split}
    \mathcal{L}_{\text{prefer}} = -\mathbb{E}_{(x, y_w, y_l) \sim \mathcal{D}} \Bigl[& \log \sigma \bigl( \beta \log \pi_{\theta}(y_w|x)\\
        &-\beta \log \pi_{\theta}(y_l|x) ) \bigr) \Bigr]
\end{split}
\end{equation*}
\begin{equation}
    \mathcal{L}_{\mathrm{CPO}}=\mathcal{L}_{\text {prefer}}+\mathcal{L}_{\mathrm{NLL}}
\label{cpo_loss}
\end{equation}
where $\pi_\theta$ is the parameterized policy, $y_w$ and $y_l$ denotes the preferred and unpreferred responses, $x$ is a set of source sentences, $\beta$ is a parameter, and $\sigma$ is the logistic function. In the next section, we assess the alignment methods, highlighting their strengths and weaknesses.

\section{Experiments}
\paragraph{Description.} 
In this section, we assess the alignment methods across three scenarios: 1) fine-tuning an SFT model with alignment methods, 2) fine-tuning a pre-trained model with alignment methods, and 3) fine-tuning an instruction-tuned model with alignment methods. Subsequently, within each scenario, we examine their performance across reasoning, mathematical problem-solving, truthfulness, question-answering, and multi-task understanding. Details regarding these scenarios are provided in the following section.

\paragraph{Evaluation Metrics.}
To evaluate the methods for reasoning, we utilize benchmarks such as ARC~\cite{clark2018think}, HellaSwag~\cite{zellers2019hellaswag}, Winogrande~\cite{sakaguchi2019winogrande}, Big Bench Sports Understanding (BB-sports), Big Bench Causal Judgment (BB-casual), Big Bench Formal Fallacies (BB-formal), and PIQA~\cite{bisk2019piqaa}. To evaluate their mathematical problem-solving abilities, we employ the GSM8K~\cite{cobbe2021training} benchmark. Truthfulness is evaluated using the TruthfulQA~\cite{lin2022truthfulqa} benchmark. Additionally, we gauge their performance in multitask understanding using the MMLU~\cite{hendrycks2021measuring} benchmark. OpenBookQA~\cite{OpenBookQA2018} and BoolQ~\cite{clark2019boolq} benchmarks are employed to assess their performance in question-answering tasks. Finally, to evaluate their effectiveness in dialog systems, we utilize MT-Bench benchmarks, which consist of 160 questions across eight knowledge domains, with GPT-4 scoring the model-generated answers on a scale from 0 to 10.
\begin{figure}[t!]
    \centering
    \includegraphics[width=7.8cm]{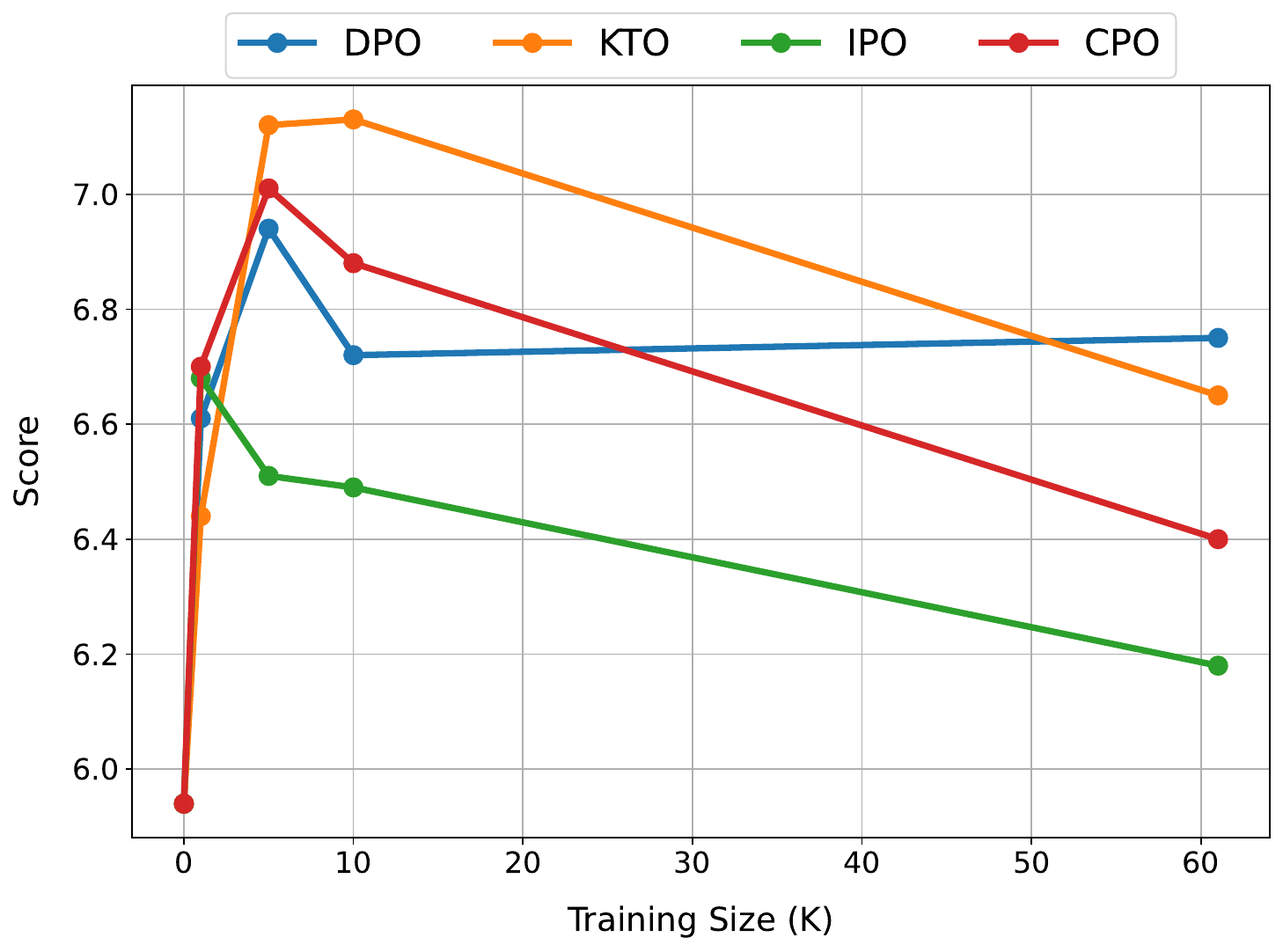}
    \caption{Comparison of performance for KTO, IPO, CPO, and DPO alignment methods on MT-Bench across various training set sizes. All methods demonstrated optimal performance with training sets ranging from 1K to 10K data points.}
    \label{fig:training_size}
\end{figure}
\subsection{Scenario 1: Fine-tune an SFT Model}
\label{sec:sc1}
\paragraph{Motivation.} 
In this scenario, we first train an SFT model and then refine it with the aforementioned alignment methods. These methods, designed to enhance the performance of DPO, have been applied to various tasks, such as machine translation. However, there hasn't been a comprehensive evaluation comparing them on the same task. The primary motivation behind these scenarios is to assess their performance across different benchmarks. Additionally, we aim to determine whether the performance of alignment methods improves with increasing training data, as it seems that alignment methods may not require extensive data beyond the SFT phase.

\paragraph{Models.}  
We employ the \texttt{zephyr-sft-full} model as our SFT model, which underwent fine-tuning utilizing the UltraChat~ \cite{ding2023enhancing} dataset. Its baseline model is \texttt{Mistral-7B-v0.1}. We proceed by training the \texttt{zephyr-sft-full} model with DPO, IPO, KTO, and CPO. For further information regarding the training and evaluation procedures, please refer to the Appendix \ref{sec:training_detain}.

\paragraph{Datasets.} 
We utilize the UltraFeedback-binarized~\cite{tunstall2023zephyr} dataset, akin to the UltraChat dataset, specifically designed for the chat completion task. Comprising 63k pairs of selected and rejected responses corresponding to specific inputs, the UltraFeedback-binarized dataset is employed for training alignment models.

\paragraph{KTO outperforms other alignment methods.}
The findings depicted in Figures \ref{fig:comparision_wo} and \ref{fig:training_size} indicate that KTO surpasses other alignment methods in MT-Bench, and across all academic benchmarks, it exhibits superior performance, with the exception of MMLU (See Table \ref{tab:mmlu}). Particularly noteworthy is KTO's remarkable performance on GSM8K, highlighting its strong aptitude for solving mathematical problems(Mathematics plot in Figure \ref{fig:comparision_wo}).

\begin{table}[h!]
    \centering
    \footnotesize
    \begin{tabular}{@{}c|ccccc@{}}
        \toprule
        \thead{Model} & \thead{DPO}& \thead{KTO}& \thead{IPO}& \thead{CPO} & \thead{SFT}
        \\
        \midrule
        Mistral & 63.14 & 62.31 & 62.44 & 62.61 & 60.92 \\
        Mistral+SFT & 59.88 & 59.53& 59.87 & 59.14 & - \\

        \bottomrule
    \end{tabular}
    \caption{Performance comparison of alignment methods on MMLU across two scenarios: 1) Fine-tuning a pre-trained model (Mistral) using alignment methods, and 2) Fine-tuning an SFT model (Mistral+SFT) using alignment methods. "-" represents that there is no value for this model. We note that the MMLU score for the Mistral model fine-tuned with SFT is 60.92.}
    \label{tab:mmlu}
\end{table}

\paragraph{Alignment methods don’t require a large training set.}
The results depicted in Figure \ref{fig:training_size} reveal that all alignment methods perform better with a smaller training set. We posit that in the typical alignment process, a significant portion of model alignment occurs during the SFT phase. Therefore, when aiming to enhance the performance of the SFT model with methods like KTO, DPO, IPO, and CPO, it is beneficial to utilize a smaller dataset for training. In essence, there exists a trade-off between aligning with SFT and aligning with RL-free methods to achieve optimal performance.

\paragraph{SFT is still enough.}
Another intriguing observation is that none of the alignment methods outperform SFT in MMLU (See Table \ref{tab:mmlu}). This suggests that SFT remains superior to other methods for multitask understanding. Additionally, apart from the KTO algorithm in reasoning, truthfulness, and question answering, SFT demonstrates comparable performance (See Reasoning, Question Answering, and Truthfulness plots in Figure \ref{fig:comparision_wo}). This indicates that alignment methods struggle to achieve notable performance improvements in these tasks.

\subsection{Scenario 2: Fine-tune a Pre-Train Model}
\label{sec:sc2}
\paragraph{Motivation.}
In this scenario, we train a pre-trained model directly with alignment methods on the UltraFeedback dataset. Several motivations underlie this scenario. Firstly, we seek to determine whether alignment methods necessitate the SFT phase. Secondly, we aim to compare the performance of models aligned with DPO, CPO, KTO, and IPO against those trained with SFT. Lastly, we aim to illustrate the impact of the SFT phase on various tasks by comparing the performance of models with and without this component.

\paragraph{Models.}
We employ \texttt{Mistral-7B-v0.1} as the pre-trained model and fine-tune it with DPO, CPO, KTO, and IPO. Further information regarding the training and evaluation process can be found in the Appendix \ref{sec:training_detain}.

\begin{table*}[h]
\centering
\scalebox{0.8}{
\begin{tabular}{c|cccccccc}
\hline
\textbf{Model} & \textbf{ARC} & \textbf{HellaSwag} & \textbf{Winogrande} & \textbf{BB-sports} & \textbf{BB-casual} & \textbf{BB-formal} & \textbf{PIQA} & \textbf{Average}

\\
\hline
Mistral-Instruct+SFT & 61.17 & 81.93 & 76.87 & 71.39 & 60 & 50.73 & 83.02 & 69.3 \\
Mistral-Instruct+IPO & 63.05 & 84.69 & 77.26 & 75.25 & 59.47 & 51.65 & 80.41 & 70.25 \\
Mistral-Instruct+KTO & 62.71 & 85.52 & 77.5 & 74.23 & 61.57 & 51.23 & 81.55 & 70.62 \\
Mistral-Instruct+CPO & 52.38 & 80.95 & 77.5 & 72.31 & 58.94 & 52.02 & 81.55 & 67.95 \\
Mistral-Instruct+DPO & 63.48 & 85.34 & 77.34 & 74.64 & 59.47 & 51.12 & 81.01 & 70.34 \\
\hline
\end{tabular}
}
\caption{Performance comparison of various alignment methods in scenario 3 on reasoning benchmarks. To assess reasoning abilities, we focused on common sense reasoning, logical reasoning, and causal reasoning (See Section~\ref{sc3}).}
\label{tab:reasoning_instruct}
\end{table*}


\begin{table*}[h]
\centering
\scalebox{0.95}{
\begin{tabular}{c|c||c||c||ccc}
\hline
\textbf{Model} & \textbf{GSM8K} & \textbf{MMLU} & \textbf{TruthfulQA} & \textbf{OpenBookQA} & \textbf{BoolQ} & \textbf{Average}

\\
\hline
Mistral-Instruct+SFT & 37.68 & 61.03 & 49.46 & 48.4 & 86.02 & 67.21 \\
Mistral-Instruct+IPO & 38.05 & 60.72 & 66.97 & 48.2 & 85.9 & 67.05  \\
Mistral-Instruct+KTO & 38.28 & 61.72 & 66.97 & 49.4 & 86.17 & 67.78 \\
Mistral-Instruct+CPO & 38.51 & 60.46 & 63.9 & 46.8 & 84.98 & 65.89 \\
Mistral-Instruct+DPO & 33.58 & 61.61 & 68.22 & 49.2 & 85.19 & 67.19 \\
\hline
\end{tabular}
}
\caption{Performance evaluation of alignment methods in scenario 3, focusing on solving mathematics problems, truthfulness, multi-task understanding, and question-answering tasks. For more detailed information, refer to Section~\ref{sc3}.}
\label{tab:math_instruct}
\end{table*}

\paragraph{Datasets.}
We train an SFT model using the UltraChat dataset, which contains 200k examples generated by \texttt{GPT-3.5-TURBO} across 30 topics and 20 text material types, providing a high-quality dataset. Additionally, for training the pre-trained model with alignment methods, we utilize the UltraFeedback dataset, as explained in Section \ref{sec:sc1}. It is worth noting that both UltraChat and UltraFeedback were curated specifically for the chat completion task.

\paragraph{KTO and CPO don’t require SFT.} 
The findings presented in Figure \ref{fig:mt_bench_with_without} indicate that skipping the SFT phase resulted in Mistral+IPO and Mistral+DPO performing poorly in the dialogue system, as they attained lower scores compared to SFT. However, Mistral+KTO and Mistral+CPO achieved scores comparable to Mistral+SFT.

\paragraph{SFT significantly affects academic benchmarks.} 
The results depicted in Figure \ref{fig:comparision_wo} reveal several key findings. Firstly, skipping the SFT phase leads to a marginal improvement in reasoning performance without significant impact. Secondly, there is a notable and consistent improvement across all alignment methods except IPO in GSM8K and TruthfulQA benchmarks. Moreover, in the MMLU benchmark, skipping the SFT phase not only enhances performance but also results in all alignment methods outperforming the SFT baseline (See Table \ref{tab:mmlu}).

\begin{table}[]
    \centering
    \footnotesize
    \scalebox{0.85}{
    \begin{tabular}{@{}cc|ccc@{}}
        \toprule
        \thead{Model} & \thead{Align} & \thead{First Turn \\ (Score)} & \thead{Second Turn \\ (Score)} & \thead{Average \\ (Score)}
        \\
        \midrule
        Mistral-Instruct & SFT & 7.78 & 7.16 & 7.47 \\
        Mistral-Instruct & DPO & 7.61 & 7.42 & 7.51 \\
        Mistral-Instruct & KTO & 7.66 & 7.36 & 7.51 \\
        Mistral-Instruct & CPO & 7.18 & 6.98 & 7.08 \\
        Mistral-Instruct & IPO & 7.88 & 7.32 & 7.60 \\

        \bottomrule
    \end{tabular}
    }
    \caption{Performance comparison of alignment methods using an instruction-tuned model without SFT on MT-Bench (More details in Section \ref{sc3}).}
    \label{tab:mt_bench_instruct}
\end{table}

\subsection{Scenario 3: Fine-tune an Instruction Tuned Model}
\label{sc3}
\paragraph{Motivation.}
The primary motivation for this scenario is to investigate the impact of the instruction-tuned model on the performance of various alignment methods. Thus, we train an instruction-tuned model with KTO, IPO, DPO, and CPO and evaluate their performance across different benchmarks. To ensure a fair comparison, we assess the performance of the alignment methods alongside the SFT method to discern their effects. Consequently, in this scenario, we bypass the SFT phase and utilize the instruction-tuned model for evaluation.

\paragraph{Models.} 
We utilize Mistral-instruct-7B-v0.2 as the instruction-tuned model and fine-tune it with DPO, CPO, KTO, and IPO. Further information regarding the training and evaluation process can be found in the Appendix \ref{sec:training_detain}.

\paragraph{Datasets.} 
Like Section \ref{sec:sc2}, we train an SFT model using the UltraChat dataset. Additionally, we employ UltraFeedback to train the pre-trained model with alignment methods, as described in scenario 1. It's worth noting that both UltraChat and UltraFeedback were curated specifically for the chat completion task.

\paragraph{Aligning an instruction-tuned model significantly affects truthfulness.}
The findings presented in Table \ref{tab:math_instruct} indicate that KTO and IPO outperform SFT by 17.5\%, whereas KTO, based on a pre-trained model, outperforms SFT by 9.5\% (See Table \ref{tab:math_without_sft} in Appendix \ref{sec:more_sc1_sc2}). This underscores the high effectiveness of an instruction-tuned model, particularly in terms of truthfulness. Additionally, it is observed that KTO surpasses other methods in MT-Bench (See Table \ref{tab:mt_bench_instruct}).

\paragraph{SFT based on instruction tuning is enough.}
The findings presented in Tables \ref{tab:reasoning_instruct} and \ref{tab:math_instruct} indicate that SFT demonstrates comparable performance across reasoning, mathematics, question-and-answer, and multi-task understanding benchmarks. While alignment methods exhibit better performance than SFT, the challenge of preparing the preference dataset remains significant, making SFT preferable in most cases. It is noteworthy that in MT-Bench, CPO performs even worse compared to SFT, suggesting that models fine-tuned with CPO exhibit weaker performance in the dialogue system compared to those fine-tuned with SFT (See Table \ref{tab:mt_bench_instruct}).

\begin{figure}[t!]
    \centering
    \includegraphics[width=7.5cm]{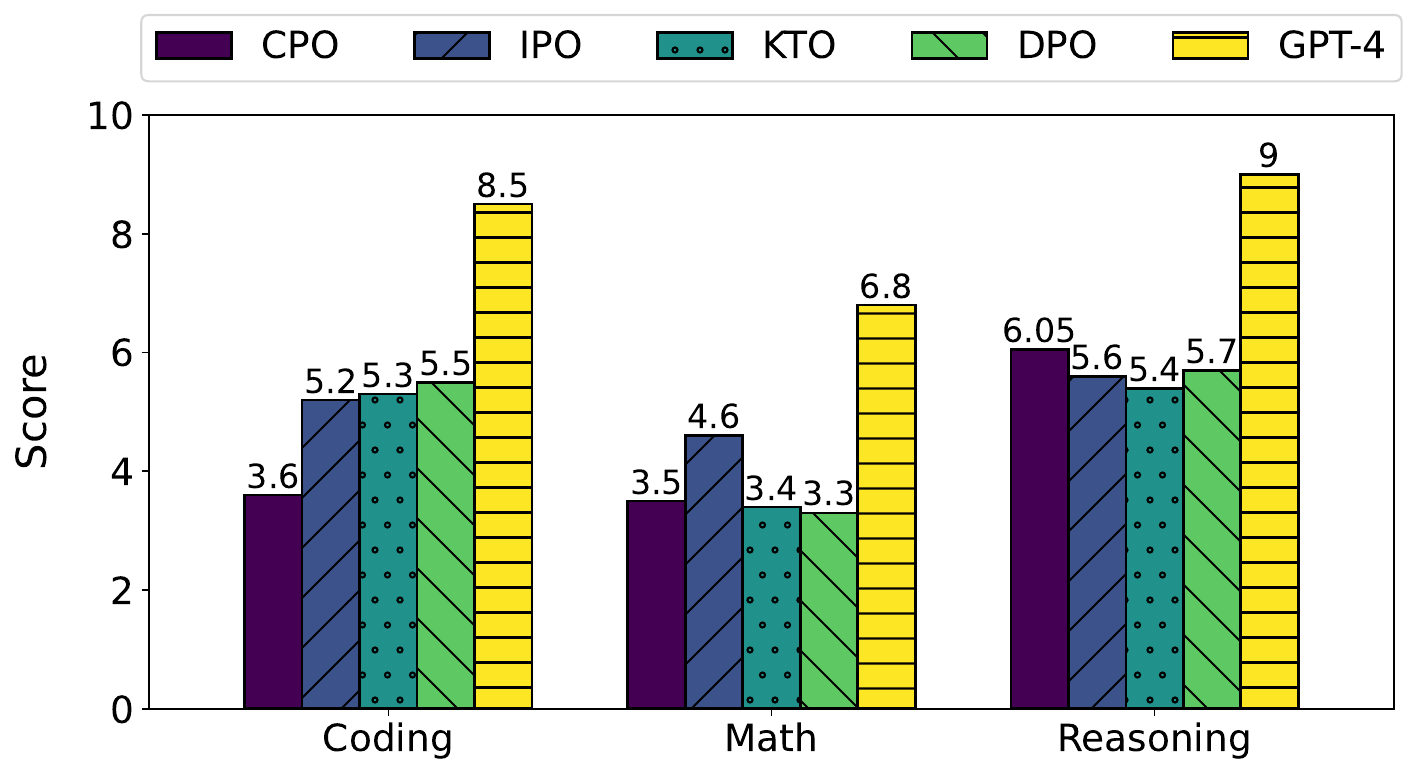}
    \caption{Performance comparison of the alignment methods based on the instruction-tuned model on MT-Bench. There exists a substantial disparity in performance between GPT-4 and alignment methods across reasoning, mathematics, and coding tasks. The score is between 0 and 10 generated by GPT-4.}
    \label{fig:weakness}
\end{figure}

\begin{figure}[t!]
    \centering
    \includegraphics[width=7.5cm]{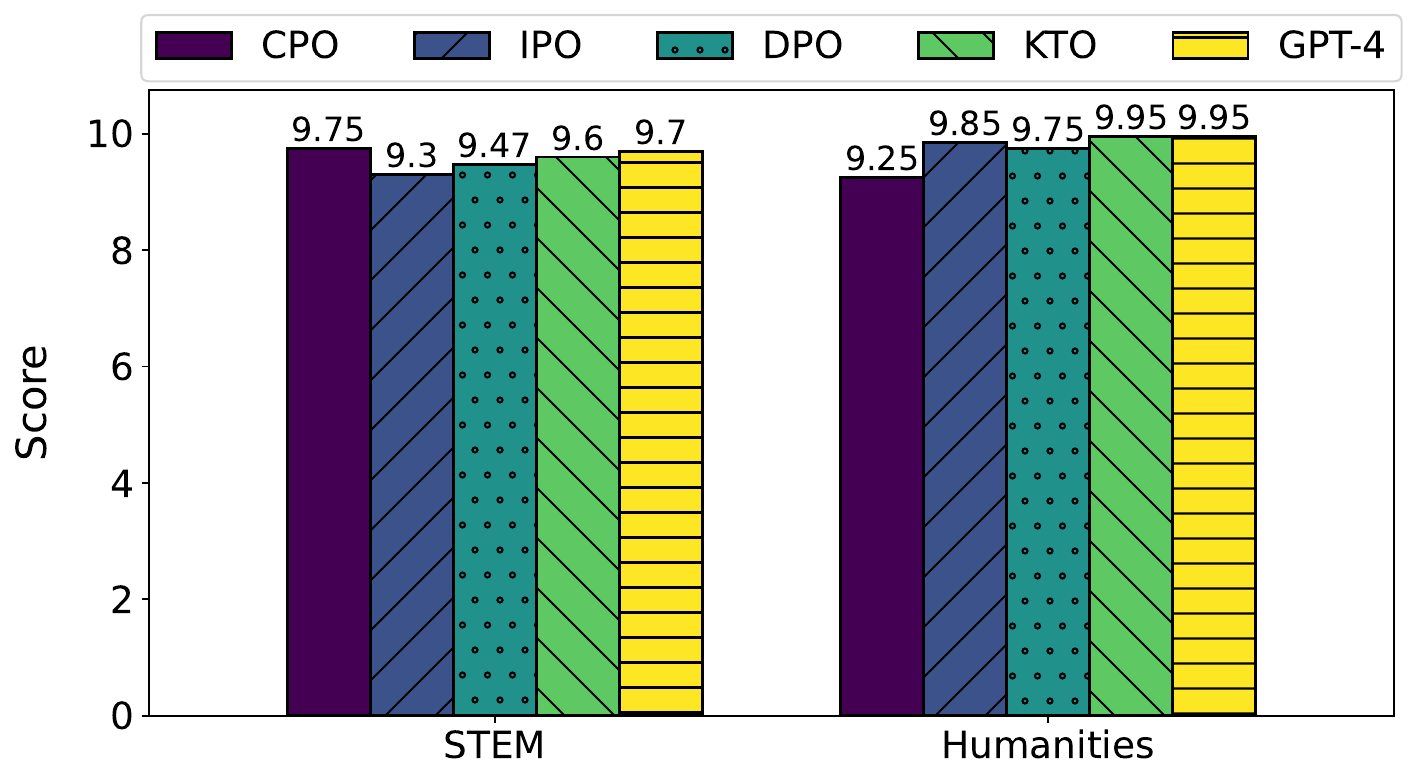}
    \caption{Alignment methods based on instruction-tuned model not only demonstrate equivalent performance to GPT-4 but can also outperform it, particularly in comparisons based on MT-Bench score. The score is between 0 and 10 generated by GPT-4.}
    \label{fig:advantage}
\end{figure}

\paragraph{Same or higher than GPT-4.} 
We observe that while improving overall performance, there is a decrease in the model's ability in certain domains (See Figure \ref{fig:weakness}). However, another intriguing discovery is that not only does KTO achieve an equal score with GPT-4 in Humanities, but CPO also outperforms GPT-4 in the STEM domain (See Figure \ref{fig:advantage}). This finding highlights the alignment methods' capability to rival state-of-the-art models such as GPT-4 with smaller models.

\section{Conclusions}
In this paper, we assessed the performance of RL-free algorithms such as DPO, KTO, IPO, and CPO across various tasks, including reasoning, mathematics problem-solving, truthfulness, question answering, and multi-task understanding in three distinct scenarios. Our findings show that KTO consistently outperforms the other alignment methods in all three scenarios. However, we noted that these techniques do not significantly enhance model performance in reasoning and question answering during regular alignment processes, though they significantly improve mathematical problem-solving. Our research also indicates that alignment methods are particularly sensitive to the volume of training data, performing best with smaller data subsets. Notably, unlike DPO, other methods, such as KTO and CPO, can bypass the SFT part and achieve comparable performance on MT-Bench. We primarily utilized an instruction-tuned model as the base for alignment, which significantly influenced truthfulness. Although this study focused on dialogue systems, we plan to extend our research to include other areas, such as safety, believing our results hold significant implications for the alignment community.


\section{Limitations} A key constraint is the challenge of preparing an appropriate dataset for training alignment methods. Furthermore, ranking multiple preferences presents another limitation that can affect the quality of the research. Inefficiencies in learning and memory also hinder progress in alignment research. Additionally, using essential benchmarks like MT-Bench and AlpacaEval \cite{dubois2023alpacafarm} is costly and necessitates access to GPT-4 for evaluation.

\section*{Ethics Statement}
We have used AI assistants (Grammarly and ChatGPT) to address the grammatical errors and rephrase the sentences.

% ...
\bibliography{anthology,custom}
\bibliographystyle{acl_natbib}

\clearpage
\appendix

\section*{Appendix}
\section{Training and Validation Details}
\label{sec:training_detain}
 We utilized the Transformer Reinforcement Learning (TRL) library for fine-tuning \cite{vonwerra2022trl}. It's noted that the notation "+" is used to indicate that a model has been fine-tuned with a specific algorithm, such as "+DPO". All models were trained using the AdamW optimizer without weight decay. Furthermore, parameter-efficient techniques such as LoRA \cite{hu2021lora} were not employed. The experiments were conducted on 6 A100 GPUs, utilizing bfloat16 precision, and typically required 5-8 hours to complete. All models are trained for one epoch, employing a linear learning rate scheduler with a peak learning rate of 5e-7 and 10\% warmup steps. Additionally, the global batch size is set to 8, and $\beta$ = 0.1 is used to regulate the deviation from the reference model. For every dataset used in our evaluation, we detail the count of few-shot examples utilized along with the specific metric employed for assessment (See Table \ref{tab:detail_benchmarks}).
 \begin{table*}[h]
\centering
\scalebox{0.6}{
\begin{tabular}{c|cccccccccccc}
\hline
\textbf{Datasets} & \textbf{ARC} & \textbf{TruthfulQA}  &  \textbf{GSM8K} &\textbf{Winogrande} & \textbf{HellaSwag} &  \textbf{MMLU} & \textbf{BB-causal} & \textbf{BB-sports} & \textbf{BB-formal} & \textbf{OpenBookQA} & \textbf{BoolQ} & \textbf{PIQA}
\\
\hline
\# few-shot & 25 & 0 & 5 & 5 & 10 & 5 & 3 & 3 & 3 & 1 & 10 & 0 \\
Metric & \texttt{acc\_norm} & \texttt{mc2} & \texttt{acc} & \texttt{acc} & \texttt{acc\_norm} & \texttt{acc} & \texttt{mc} & \texttt{mc} & \texttt{mc} &  \texttt{acc\_norm} & \texttt{acc} & \texttt{acc\_norm} \\

        \bottomrule
    \end{tabular} }
    \caption{Detailed information of Open LLM Leaderboard, Big Bench and other benchmarks.}
    \label{tab:detail_benchmarks}
\end{table*}

\section{More Details for Scenarios 1 and 2}
\label{sec:more_sc1_sc2}
In this section, we present the details for reasoning benchmarks for scenario 1 in Table \ref{tab:reasoning_sft} and for scenario 2 in Table \ref{tab:reasoning_without_sft}. Additionally, we provide details for other benchmarks in Tables \ref{tab:math_sft} and \ref{tab:math_without_sft}.

\begin{table*}[h]
\centering
\scalebox{0.8}{
\begin{tabular}{c|cccccccc}
\hline
\textbf{Model} & \textbf{ARC} & \textbf{HellaSwag} & \textbf{Winogrande} & \textbf{BB-sports} & \textbf{BB-casual} & \textbf{BB-formal} & \textbf{PIQA} & \textbf{Average}

\\
\hline
Mistral+SFT     &60.41 & 81.69 & 74.19 & 61.76 & 51.57 & 51.4 & 81.66 &  66.09\\
Mistral+SFT+DPO &61.60 & 82.11 & 77.82 & 72.31 & 51.57 & 51.28 & 81.33 &  65.64\\
Mistral+SFT+IPO &59.56 & 81.08 & 76.55 & 68.76 & 51.05 & 52.03 & 81.55 &  67.22\\
Mistral+SFT+CPO &54.52 & 79.24 & 76.4 & 72.21 & 53.68 & 52.18 & 80.9 &  67.1\\
Mistral+SFT+KTO &57.84 & 82.19 & 77.26 & 73.52 & 57.89 & 51.19 & 81.93 &  68.83\\
\hline
\end{tabular}
}
\caption{Performance comparison of the various alignment methods in scenario 1 on reasoning benchmarks. To assess reasoning abilities, we focused on common sense reasoning, logical reasoning, and causal reasoning.}
\label{tab:reasoning_sft}
\end{table*}



\begin{table*}[h]
\centering
\scalebox{0.8}{
\begin{tabular}{c|cccccccc}
\hline
\textbf{Model} & \textbf{ARC} & \textbf{HellaSwag} & \textbf{Winogrande} & \textbf{BB-sports} & \textbf{BB-casual} & \textbf{BB-formal} & \textbf{PIQA} & \textbf{Average}

\\
\hline
Mistral+SFT & 60.41 & 81.69 & 74.19 & 61.76 & 51.57 & 51.4 & 81.66 & 66.09\\
Mistral+DPO & 63.82 & 84.99 & 78.92 & 74.64 & 57.89 & 50.69 & 83.02 & 70.56 \\
Mistral+IPO & 68 & 81.7 & 77.03 & 73.93 & 58.94 & 52.3 & 83.18 & 70.72 \\
Mistral+CPO & 60.49 & 82.21 & 78.45 & 72 & 55.78 & 52.88 & 82.15 & 69.13\\
Mistral+KTO & 64.5 & 85.31 & 78.68 & 77.68 & 56.84 & 51.05 & 83.35 & 71.05 \\
\hline
\end{tabular}
}
\caption{Performance comparison of the various alignment methods in scenario 2 on reasoning benchmarks. To assess reasoning abilities, we focused on common sense reasoning, logical reasoning, and causal reasoning.}
\label{tab:reasoning_without_sft}
\end{table*}



\begin{table*}[h]
\centering
\scalebox{0.95}{
\begin{tabular}{c|c||c||c||ccc}
\hline
\textbf{Model} & \textbf{GSM8K} & \textbf{MMLU} & \textbf{TruthfulQA} & \textbf{OpenBookQA} & \textbf{BoolQ} & \textbf{Average}

\\
\hline
Mistral+SFT     & 26.76 & 60.92 & 43.73 & 43.2 & 85.16 & 64.18 \\
Mistral+SFT+DPO & 30.62 & 59.88 & 44.78 & 46 & 85.29 &  65.64 \\
Mistral+SFT+IPO & 31.31 & 59.87 & 41.37 & 45 & 84.77 & 64.88 \\
Mistral+SFT+CPO & 27.89 & 59.14 & 45.1 & 44 & 84.28 & 64.14 \\
Mistral+SFT+KTO & 34.72 & 59.53 & 45.9 & 47 & 85.87 & 66.43 \\
\hline
\end{tabular}
}
\caption{Evaluation of alignment methods in scenario 1, focusing on solving mathematics problems, truthfulness,  multi-task understanding, and question-answering tasks.}
\label{tab:math_sft}
\end{table*}
\begin{table*}[h]
\centering
\scalebox{0.95}{
\begin{tabular}{c|c||c||c||ccc}
\hline
\textbf{Model} & \textbf{GSM8K} & \textbf{MMLU} & \textbf{TruthfulQA} & \textbf{OpenBookQA} & \textbf{BoolQ} & \textbf{Average}

\\
\hline
Mistral+SFT & 26.76 & 60.92 & 43.73 & 43.2 & 85.16 & 64.18 \\
Mistral+DPO & 36.01 & 63.14 & 51.2 & 49.4 & 86.78 & 68.09  \\
Mistral+IPO & 19.86 & 62.44 & 52.28 & 50 & 86.78 & 68.39 \\
Mistral+CPO & 34.19 & 62.61 & 50.04 & 47.4 & 86.14 & 66.77 \\
Mistral+KTO & 42.15 & 62.31 & 52.98 & 48.8 & 86.78 & 67.79 \\
\hline
\end{tabular}
}
\caption{Evaluation of alignment methods in scenario 2, focusing on solving mathematics problems, truthfulness,  multi-task understanding, and question-answering tasks.}
\label{tab:math_without_sft}
\end{table*}

\end{document}